\documentclass[sigconf]{acmart}

\AtBeginDocument{
    
}
\usepackage{float}
\usepackage{caption}
\usepackage{enumitem}
\usepackage{algorithm}
\usepackage{algpseudocode}
\usepackage{amsmath}
\usepackage{graphicx}
\usepackage{multirow}
\usepackage{placeins}
\usepackage{siunitx}
\usepackage{tikz}
\usepackage{hyperref}
\hypersetup{
    colorlinks=true,
    linkcolor=blue,
    filecolor=magenta,      
    urlcolor=cyan,
}
\usetikzlibrary{arrows.meta, calc, fit, positioning, shapes.geometric}

\begin{document}

%% The "title" command has an optional parameter, allowing the author to define a "short title" to be used in page headers.
\title{ReProCon: Scalable and Resource-Efficient Few-Shot Biomedical Named Entity Recognition}

\author{Jeongkyun Yoo}
\affiliation{
    \institution{Ain Hospital}
    \city{Incheon}
    \country{South Korea}}
\email{luxiante@gmail.com}

\author{Nela Riddle}
\affiliation{
    \institution{Indiana University}
    \city{Indianapolis}
    \country{USA}}
\email{nmriddle@iu.edu}

\author{Andrew Hoblitzell}
\affiliation{
    \institution{Purdue University}
    \city{Indianapolis}
    \country{USA}}
\email{ahoblitz@purdue.edu}
\date{August 2025}

%% This command allows the author to define a more concise list of authors' names for this purpose.
\renewcommand{\shortauthors}{Yoo, Riddle, and Hoblitzell}

%% The abstract
\begin{abstract}
Named Entity Recognition (NER) in biomedical domains faces challenges due to data scarcity and imbalanced label distributions, especially with fine-grained entity types. We propose ReProCon, a novel few-shot NER framework that combines multi-prototype modeling, cosine-contrastive learning, and Reptile meta-learning to tackle these issues. By representing each category with multiple prototypes, ReProCon captures semantic variability, such as synonyms and contextual differences, while a cosine-contrastive objective ensures strong interclass separation. Reptile meta-updates enable quick adaptation with little data. Using a lightweight fastText + BiLSTM encoder, with much lower memory use, ReProCon achieves a macro-F\textsubscript{1} score close to BERT-based baselines ($\sim$99\% of BERT performance). The model remains stable with a label budget of 30\% and only drops 7.8\% in F\textsubscript{1} when expanding from 19 to 50 categories, outperforming baselines such as SpanProto and CONTaiNER, which see 10–32\% degradation in Few-NERD. Ablation studies highlight the importance of multi-prototype modeling and contrastive learning in managing class imbalance. Despite difficulties with label ambiguity, ReProCon demonstrates state-of-the-art performance in resource-limited settings, making it suitable for biomedical applications. \\

%% ArXiv licence (CC4)
\small
\noindent
\textbf{Disclaimer: This work is licensed under a Creative Commons Attribution 4.0 International License (CC BY 4.0).} \\
%% Correspondence
\textbf{Correspondence: Nela Riddle (nmriddle@iu.edu), Andrew Hoblitzell (ahoblitz@purdue.edu)}
\end{abstract}

%% This command processes the author, affiliation, and title information and builds the first part of the formatted document.
\maketitle

%% The code below is generated by the tool at http://dl.acm.org/ccs.cfm. Please copy and paste the code instead of the example below.

%% Separate the keywords with commas.

% \received{20 February 2025}
% \received[revised]{12 March 2025}
% \received[accepted]{5 June 2025}

\section{Introduction}
\label{sec:introduction}
Named Entity Recognition (NER) identifies and categorizes entity mentions in unstructured text, serving as a foundation for downstream biomedical NLP tasks, including knowledge graph construction, literature curation, and clinical decision support. In biomedical domains, NER is particularly challenging due to the high cost of annotation and the long-tailed label distributions in public corpora, such as MedMentions~\cite{sunil2019medmentions}, BC2GM~\cite{krallinger2008biocreative}, and NCBI-Disease~\cite{dougan2014ncbi}, which often contain contextually variable entity categories.
    
Few-shot learning mitigates the data bottleneck by training on only a handful of labeled examples. Although recent work shows promise on general domain benchmarks (e.g., Few-NERD~\cite{ding2021fewnerd}), NER has achieved limited success in the biomedical domain with a large number of entity types. In such cases, existing approaches often suffer from poor performance due to the overwhelming diversity of categories and a lack of ontological awareness.

To address these challenges, we propose \textbf{ReProCon}, a prototype-based cosine-contrastive NER framework that addresses these gaps.
The key components are the following:
\begin{enumerate}[leftmargin=*,itemsep=0.3em]
    \item \textbf{Multi-prototype modeling}: Each category has $M$ distinct prototypes that capture various contexts of biomedical entities, designed to be angularly separated for better differentiation.
    
    \item \textbf{Cosine supervised-contrastive objective}: The optimization aims for the entity spans to align closely with their corresponding category prototypes while being angularly distanced from others.
    
    \item \textbf{First-order meta-updates (Reptile)}: Episode-based updates enable rapid parameter adaptation without the computational burden of second-order gradients~\cite{nichol2018reptile}.
\end{enumerate}

Empirical evaluation in MedMentions demonstrates that ReProCon effectively tackles imbalanced samples and hierarchical complexity of the dataset, such as concept structures based on the Unified Medical Language System (UMLS)~\cite{bodenreider2004umls}, while achieving performance comparable to or exceeding state-of-the-art methods, such as BioBERT~\cite{lee2020BioBERT} and SpanProto~\cite{wang2022spanproto}, with improved computational efficiency.

\section{Related Work}
\label{sec:related_works}
\subsection{Domain-specific Language Models and Biomedical Resources}
\label{ssec:domain_specific_language_models_and_biomedical_resources}
\paragraph{Biomedical PLMs.}
Large-scale pre-training on domain-specific corpora significantly enhances biomedical NLP performance. Models like BioBERT, PubMedBERT~\cite{gu2021domain}, and SciBERT~\cite{beltagy2019scibert} utilize PMC articles or PubMed abstracts instead of generic BERT training data, achieving substantial improvements in biomedical NER and question answering. These results underscore the data mismatch, differences in vocabulary, and context between the general and biomedical domains as the main bottleneck for domain transfer.

\paragraph{UMLS.}
The Unified Medical Language System (UMLS) integrates more than 100 biomedical vocabularies into a unified ontology. Despite its extensive coverage, UMLS's heterogeneous sources, biannual updates, and complex licensing pose challenges for few-shot evaluations, particularly in data preparation and model adaptation. To address this, we derive a dynamic semantic type hierarchy from MedMentions annotations~\cite{sunil2019medmentions}, enabling ReProCon to capture the latest ontological structures while avoiding licensing constraints.

\subsection{Prototypical Network}
\label{ssec:prototypical_network}
Several studies have implemented the prototypical network for the NER few-shot learning task. Classic prototypical networks assume one centroid per category~\cite{snell2017prototypical}. ReProCon extends hyperspherical prototype networks by incorporating multiple prototypes per category, enabling a finer-grained representation of biomedical entities.

Mettes et al. applied hyperspherical prototype networks to NER tasks, including those within and between domains. They adapted classic prototypical networks by training prototypes as part of the end-to-end process, using a loss term designed to maximize the separation between prototypes~\cite{mettes2019hyperspherical}. In this study, we adopted the concept of calculating the repulsion loss between prototypes.

\subsection{Meta-Learning}
\label{ssec:meta_learning}
Model-Agnostic Meta-Learning learns a single parameter initialization that can fine‑tune new tasks with only a few gradient updates~\cite{finn2017maml}. It employs an inner loop for task-specific fine-tuning within an outer loop for generalized parameter optimization that computes second-order gradients. This approach aims to produce model parameters that are adaptable to new tasks with minimal data. However, backpropagating through gradient steps introduces significant computational and memory overhead, often causing instability in deep or noisy tasks.

Reptile is a first-order meta-learning algorithm that learns a task‑agnostic parameter initialization by repeatedly fine-tuning the sampled tasks and adjusting the shared weights to the adapted ones. Unlike MAML, Reptile eliminates second-order derivatives, so its meta-update is both simpler and substantially faster, while retaining strong few-shot performance. This outer loop translation of inner loop progress approximates higher-order information.

\subsection{Supervised Contrastive Learning}
\label{ssec:supervised_contrastive_learning}
Supervised contrastive learning improves representation robustness in a few-shot setting and has also been applied in recent NER tasks to improve representation robustness under limited supervision~\cite{khosla2021supervised}. For example, Das et al. implemented contrastive learning for the NER task by calculating losses over entity spans~\cite{das2022container}. However, existing formulations assume one-to-one prototype alignment, which struggles with fine-grained biomedical categories. We extended this approach by introducing a multi-prototype contrastive loss that uses category-wise minimum pooling to align projected spans with the closest prototype in each category. This loss function enables both semantic flexibility and interclass separation, which is crucial for fine-grained classification of biomedical entities.

\section{Data Processing}
\label{sec:data_processing}
\subsection{Sentence and Entity Preprocessing}
\label{ssec:sentence_and_entity_preprocessing}
We first normalize whitespace and remove text enclosed in parentheses, brackets, braces, or angle brackets, as they often duplicate named entities, introducing redundant contextual variability. We applied the same preprocessing to both the sentences and their entity strings.

\subsection{Semantic Type Normalization and Hierarchical Disambiguation}
\label{ssec:semantic_type_normalization_and_hierarchical_disambiguation}
The MedMentions corpus annotates entities with semantic type IDs of the Unified Medical Language System (UMLS), many of which are fine-grained (e.g., \verb | Organization |, \verb | Substance |), exhibiting highly imbalanced sample distributions. Moreover, a single span can carry multiple types of labels, introducing ambiguity. To enhance learning efficiency by focusing on well-represented categories, we design a two-stage procedure through sparsity reduction and PageRank-based disambiguation to select a single type when multiple labels are present.

\subsubsection{Category Integration and Pruning}
\label{ssec:category_integration_and_pruning}
We map every semantic type ID to a human-readable name using a lookup table provided from the Medmentions dataset. Let $Y$ denote the original set of types and $\mathrm{freq}(y)$ the frequency of type $y\in{Y}$. Then, to reduce sparsity and avoid underrepresented categories, two rules are applied:

\begin{enumerate}[leftmargin=*,itemsep=0.3em]
    \item \textbf{Depth constraint} If a semantic type ID corresponds to a node deeper than a specific level in the UMLS hierarchy, we promote it to its closest ancestor. We tested two constraint levels, levels 3 and 4, to compare the model's performance according to the number of categories.
    
    \item \textbf{Low-frequency merging} If a type has fewer than a specific threshold number of samples, it is recursively promoted to the closest ancestor until the sample number becomes larger than the threshold, regardless of its depth. We set thresholds of 100 for level 3 constraints and 50 for level 4 constraints. 
\end{enumerate}

After applying the above rules, we discarded any top-level category with fewer than the threshold number of instances. The resulting taxonomies yield 19- and 50-way classification settings; the list of category statistics is in Table~\ref{tab:19categories} and Table~\ref{tab:50categories} (Appendix).

\subsubsection{Resolution of Multi-Label Entities}
\label{ssec:resolution_of_multi_label_entities}
As described in Algorithm~\ref{alg:pagerank_hierarchy}, we constructed an undirected co-occurrence graph where nodes represent semantic types, with edges reflecting the frequency of co-occurrence within entity annotations. We then used PageRank centrality to rank the types, assuming that the most central types are broader or more representative categories~\cite{brin1998pagerank}. For entities with multiple semantic types, we retained only the one with the highest PageRank score, reducing the labeling noise.

\subsection{Span Generation and Label Assignment}
\label{ssec:span_generation_and_label_assignment}
We used SpaCy~\cite{honnibal2020spacy} to process tokenization before generating text spans. Inspired by EPNet~\cite{ji2022fewshot}, we generated text spans from sentence token sequences $S = [{t_1}, {t_2}, ..., {t_n}]$ by extracting consecutive tokens of up to 8 tokens, since 99.95\% of the entity spans contain fewer than nine tokens. To preserve contextual order, we create samples by marking the generated token spans with a special marker \verb| [MARK_POSITION] | within the sentences to disambiguate multiple instances of the same word inside a sentence. We organized these samples by semantic type category.

\begin{algorithm}[H]
\caption{PageRank-based Construction of a Semantic-Type Hierarchy}
\label{alg:pagerank_hierarchy}
\begin{algorithmic}[1]
\Require Annotated corpus $D=\{d_1,\dots,d_{|D|}\}$ with entity spans
\Ensure Ordered list $H$ of semantic types ranked by PageRank

\State $\mathcal{C} \leftarrow [\,]$        \Comment{list of multi type combinations}
\State $V \leftarrow \varnothing$           \Comment{set of all unique semantic type IDs}

\ForAll{document $d\in D$}
    \ForAll{entity span $e\in d$}
        \State $Y \gets \{y \in e \mid \mathrm{freq}(y) \geq \text{threshold}\}$   \Comment{Filter types below frequency threshold}
        \If{$|Y|>1$ \textbf{and} $Y\notin \mathcal{C}$}
            \State append $Y$ to $\mathcal{C}$
            \State $V \leftarrow V \cup Y$
        \EndIf
    \EndFor
\EndFor

\State $A \leftarrow \mathbf{0}_{|V|\times|V|}$     \Comment{undirected co-occurrence matrix}

\ForAll{$Y \in \mathcal{C}$}
    \ForAll{$(y_i,y_j)\in\text{Combinations}(Y,2)$}     \Comment{pairwise type combinations}
        \State $A[y_i,y_j] \leftarrow A[y_i,y_j] + 1$
        \State $A[y_j,y_i] \leftarrow A[y_j,y_i] + 1$   \Comment{symmetry}
    \EndFor
\EndFor

\State Build graph $G=(V,E)$ with edge weights $A[y_i,y_j]$
\State $\text{PR} \gets \textsc{PageRank}(G)$           \Comment{damping factor $0.85$}

\State $H \gets$ types in $V$ sorted by $\text{PR}[y]$ in descending order
\State \Return $H$
\end{algorithmic}
\end{algorithm}

\subsection{Task Sampling Strategy for Reptile Meta-Learning}
\label{ssec:task_sampling_strategy_for_reptile_meta_learning}
Let $K$ denote the number of support shots and define the split ratio of category-wise training as $r_{\text{train}}\in\{0.3,0.4,\dots,0.8\}$. \footnote{Validation and query splits are fixed at 10\% and $(1-r_{\text{train}})-0.10$, respectively.} For each episode, we:

\begin{enumerate}[leftmargin=*,itemsep=0.3em]
    \item sample $N$ (either 19 or 50 in this study) categories uniformly,
    
    \item draw $K$ support examples per category from the training pool,
    
    \item draw validation and query examples from their respective pools.
\end{enumerate}

To control memory usage, we limit pool sizes to 30,000 (support), 500 (validation), and 400 (query) instances per category and pregenerate 200 episodic task sets. The learning process performs inner loop updates followed by a meta-update with step size $\alpha$.

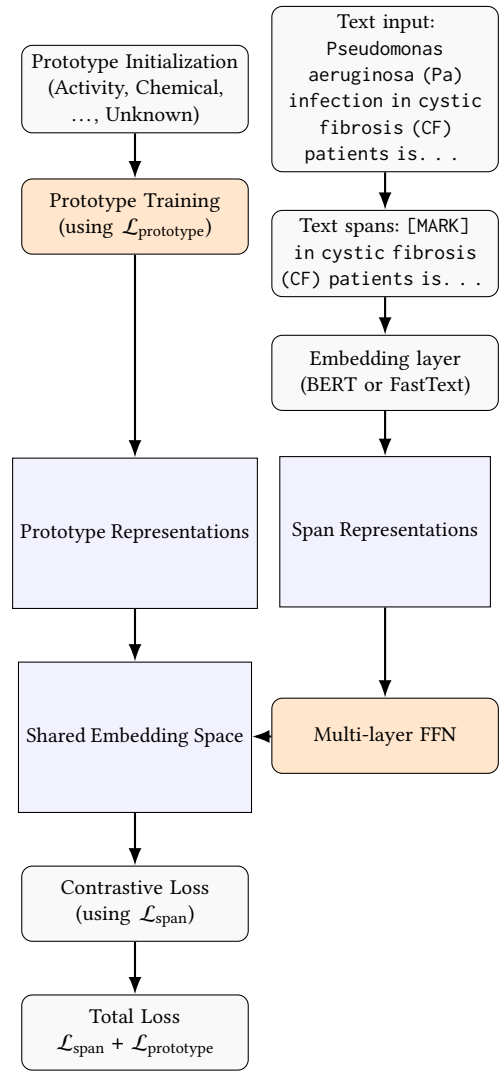
\begin{figure}[ht]
    \centering
\begin{tikzpicture}[
    font=\small,
    box/.style={draw, rounded corners, fill=gray!5, text width=2.8cm, align=center, minimum height=1cm},
    arrow/.style={-Latex, thick},
    box3d/.style={draw, minimum width=2.8cm, minimum height=2cm, fill=blue!5, align=center},
    param/.style={draw, rounded corners, fill=orange!20, text width=2.8cm, align=center, minimum height=1cm}
]

% Step 1
\node[box] (init) {Prototype Initialization\\(Activity, Chemical, \ldots, Unknown)};
\node[param, below=0.6cm of init] (training) {Prototype Training\\(using $\mathcal{L}_{\text{prototype}}$)};

% Step 4-7 (text spans to BERT to span reps)
\node[box, right=0.3cm of init] (input) {Text input: \texttt{Pseudomonas aeruginosa (Pa) infection in cystic fibrosis (CF) patients is\ldots}};
\node[box, below=0.5cm of input] (spans) {Text spans: \texttt{[MARK] in cystic fibrosis (CF) patients is\ldots}};
\node[box, below=0.5cm of spans] (bert) {Embedding layer (BERT or FastText)};
\node[box3d, below=0.6cm of bert] (span3d) {Span Representations};

\node[box3d, below=2.7cm of training] (proto3d) {Prototype Representations};

% Step 8: FFN
\node[param, below=1.2cm of span3d] (ffn) {Multi-layer FFN};

% Step 9
\node[box3d, below=0.7cm of proto3d] (sharedspace) {Shared Embedding Space};

% Step 10
\node[box, below=0.7cm of sharedspace] (match) {Contrastive Loss\\(using $\mathcal{L}_{\text{span}}$)};

% New box for total loss
\node[box, below=0.7cm of match] (totalloss) {Total Loss $\mathcal{L}_{\text{span}}+\mathcal{L}_{\text{prototype}}$};

% Arrows
\draw[arrow] (init) -- (training);
\draw[arrow] (training) -- (proto3d);
\draw[arrow] (proto3d) -- (sharedspace);
\draw[arrow] (span3d) -- (ffn);
\draw[arrow] (ffn) -- (sharedspace);
\draw[arrow] (sharedspace) -- (match);

\draw[arrow] (input) -- (spans);
\draw[arrow] (spans) -- (bert);
\draw[arrow] (bert) -- (span3d);
\draw[arrow] (match) -- (totalloss);

\end{tikzpicture}

\caption{A high-level overview of the model pipeline for a single task, where orange boxes indicate learnable parameters. The system begins with the initialization and training of the prototype. Text inputs are tokenized into spans, embedded (e.g., via BERT~\cite{devlin2019bert}), and transformed into span representations.}
    
\Description{Diagram showing the model pipeline: initialization, training, text input, embedding, span representations, feed-forward network, shared embedding space, and loss calculation.}
\label{fig:pipeline_overview}
\end{figure}

\section{Model Training}
\label{sec:model_training}
\subsubsection*{Notation}
Let $N$ be the number of semantic categories, $M$ the number of prototypes per category, and $D=50$ the dimensionality of each prototype. The shared span-representation space has dimensionality $d_{\text{representation}}=512$. For the fastText branch, we use static embeddings of size $d_{\text{embed}}=300$ and sinusoidal positional encodings of size $d_{\text{position}}=200$. We initialized all weights with Xavier Uniform~\cite{glorot2010understanding} under a fixed seed 42 for reproducibility. We ran all experiments on a single NVIDIA L4 GPU (22.5GB VRAM) within Google Colab's high-RAM environment. \footnote{All codes and experiment results are in the following GitHub repository: \url{https://github.com/FewshotMedicalNer/fewshot_medical_ner}}

\subsection{Prototype Matrix}
\label{ssec:prototype_matrix}
Each category maintains $M=10$ learnable prototypes to capture semantic diversity. The resulting tensor is
\[
    \mathbf{P}\in\mathbb{R}^{(N\times M)\times D}.
\]
Every prototype vector is $\ell_2$ normalized after each meta-update to calculate cosine similarity.

\subsection{Context Encoder (fastText Setting)}
\label{ssec:context_encoder_fasttext_setting}
Given a token sequence of length $T\!=\!300$, we concatenate static fastText embeddings\cite{bojanowski2017fasttext} with positional encodings, described as
\[
    \mathbf{X}\in\mathbb{R}^{T\times(d_{\text{embed}}+d_{\text{position}})}.
\]
A single-layer bidirectional LSTM\cite{graves2005bilstm} with hidden size $h=1024$ per direction produces contextual states:
\[
    \mathbf{H}
    =\mathrm{BiLSTM}(\mathbf{X})
    \in\mathbb{R}^{T\times 2h}.
\]
A linear projection $\mathbf{W}_p\in\mathbb{R}^{2h\times d_{\text{representation}}}$ projects token representations in the $d_{\text{representation}}$ space, to align with the output of the BERT encoder in our alternative configuration.

\subsection{Sinusoidal Positional Encoding}
\label{ssec:sinusoidal_positional_encoding}
We adopt the sinusoidal encoding from \cite{vaswani2017attention}. For positions $p\in[0,T-1]$ and dimensions $i\in \bigl[0,\frac{d_{\text{pos}}}{2}-1 \bigr]$
\begin{align}
P_{p,\,2i}   &= \sin\!\bigl(p/10000^{2i/d_{\text{position}}}\bigr),\\
P_{p,\,2i+1} &= \cos\!\bigl(p/10000^{2i/d_{\text{position}}}\bigr),
\end{align}
The matrix $\mathbf{P}\in\mathbb{R}^{T\times d_{\text{position}}}$ is concatenated with $\mathbf{X}$ along the feature dimension.

\subsection{Span Projection Network}
\label{ssec:span_projection_network}
The span embedding of each candidate $\mathbf{s}\in\mathbb{R}^{d_{\text{representation}}}$ is processed by a feed-forward network of $L$ layers:
\[
    \mathbf{v}^{(\ell+1)}
    =\mathrm{Dropout}\!\bigl(
        \mathrm{GELU}\!\bigl(
            \mathrm{BatchNorm}\bigl(\mathbf{W}^{(\ell)}\mathbf{v}^{(\ell)}+\mathbf{b}^{(\ell)}\bigr)
        \bigr)
    \bigr),
    \
    \mathbf{v}^{(0)}=\mathbf{s}.
\]
A final affine transformation followed by Layer Normalization~\cite{ba2016layer} projects into the prototype space:
\[
    \mathbf{z}
    =\mathrm{LayerNorm}\!\bigl(
        \mathbf{W}^{(\text{out})}\mathbf{v}^{(L)}+\mathbf{b}^{(\text{out})}
    \bigr)
    \in\mathbb{R}^{D}.
\]
We computed cosine similarity between the $\mathbf{z}$ and prototype vectors during training and inference. In this study, we set $L=1$ to avoid overfitting.

\subsection{Span Representation Embedding}
\label{ssec:span_representation_embedding}
We evaluate two span-encoding strategies:
\begin{enumerate}[leftmargin=*,itemsep=0.3em]
    \item \textbf{fastText + BiLSTM branch} For each SpaCy token, we obtain a fastText vector $\mathbf{e}\in\mathbb{R}^{d_{\text{embed}}}$ and apply the weighted mean for multi-word tokens:
    \[
        \mathbf{e}
        = \sum_{i=0}^{C-1} w_i\,\hat{\mathbf{e}}_i,
        \quad
        w_i = 1 + 0.1\!\left(\tfrac{C-i}{2}\right)^{\!2},
        \quad
        \hat{\mathbf{e}}_i = \frac{\mathbf{e}_i}{\lVert\mathbf{e}_i\rVert_2}.
    \]
    The token sequence is right-padded to length~$T$ with zeros, and concatenated with positional encodings:
    \[
        \mathbf{X}
        =\bigl[\mathbf{E}\;\Vert\;\mathbf{P}\bigr]
        \in\mathbb{R}^{T\times(d_{\text{embed}}+d_{\text{position}})}.
    \]
    A single layer BiLSTM ($h{=}1024$ per direction) produces contextual states $\mathbf{H}\in\mathbb{R}^{T\times2h}$. The hidden state at the marker position becomes the span representation $\mathbf{s}\in\mathbb{R}^{d_{\text{representation}}}$.
    
    \item \textbf{BERT branch} Let $(s_{\text{start}}, s_{\text{end}})$ denote the range of subtoken indexes of the marked span (computed by aligning SpaCy tokens with the BERT wordpiece sequence and skipping all [CLS]/[SEP] markers). We feed the entire sequence into the vanilla BERT-base-cased model and obtain hidden states $\mathbf{H}\in\mathbb{R}^{T\times d_{\text{bert}}}$. The span representation $\mathbf{s}\in\mathbb{R}^{d_{\text{representation}}}$ is a pooled tensor, using the mean-max scheme. As BERT inherently captures bidirectional context, this branch does not need an extra BiLSTM layer.
\end{enumerate}

\subsection{Loss Functions}
\label{ssec:loss_functions}
The total objective is the sum of a \emph{prototype repulsion loss} term and a \emph{span alignment loss} term:
\[
    \mathcal{L}
    =\mathcal{L}_{\text{proto}}+\mathcal{L}_{\text{span}}.
\]

\paragraph{Prototype repulsion loss.}
Inspired by a methodology suggested in EPNet, we maximize the angular separation between $N\times M$ prototypes
$\{\mathbf{p}_0,\dots,\mathbf{p}_{N\times M-1}\}$:
\[
    \mathcal{L}_{\text{proto}}
    =\frac{1}{N M}
    \sum_{i=0}^{N M-1}
    \max_{j\ne i}
    \bigl(\cos(\mathbf{p}_i,\mathbf{p}_j)+1\bigr).
\]
The offset $+1$ ensures a positive quantity and penalizes highly aligned pairs.

\paragraph{Span alignment loss.}
Let $\{\mathbf{q}_{0},\dots,\mathbf{q}_{NK-1}\}$ be the projected span vectors of a training episode ($K$ spans per category). We first compute the squared angular distance $H_{ij}=(1-\cos(\mathbf{p}_i,\mathbf{q}_j))^{2}$.
For each category $c$, we take the minimum distance between its $M$ prototypes:
\[
    \widetilde{H}_{c,j}
    =\min_{cM\le k<(c+1)M} H_{k j}.
\]
The span alignment loss encourages each query span to align closely with the prototype of its category while remaining distant from the others, enabling a robust classification of biomedical entities:
\[
    \mathcal{L}_{\text{span}}
    =\sum_{c=0}^{N-1}
        \frac{\;\frac1K\sum_{j=cK}^{(c+1)K-1}\widetilde{H}_{c,j}\;}
        {\sum_{j=0}^{NK-1}\widetilde{H}_{c,j}}.
\]

\subsection{Meta-Learning with Reptile}
\label{ssec:meta_learning_with_reptile}
Each meta-episode involves (1) sampling a few-shot task from the pool, (2) fine-tuning a task model in its support set under the contrastive objective of Section~\ref{ssec:loss_functions}, and (3) performing a first-order meta-learning update to the meta model. The algorithm~\ref{alg:meta_training} describes the procedure.

We set inner epochs to $E=5$ (fastText) and $E=3$ (BERT), with learning rates $\gamma=5\times10^{-4}$ and $\gamma=10^{-3}$, respectively, optimized by cross-validation. The meta-step sizes are $\alpha=0.4$ and $\alpha=0.5$. Cosine decay scheduling~\cite{loshchilov2017sgdr} and gradient clipping~\cite{pascanu2013difficulty} (max norm 1.0) stabilize the training.

\subsection{Hard-Negative Training}
\label{ssec:hard_negative_training}
After the initial meta-training phase, we identify hard-negative spans that the model assigns high scores to despite belonging to other categories. We conducted a second meta-training round using the resampled support sets under the same hyperparameters. We compared the performance of the model against a model trained without hard-negative mining to evaluate changes in classification accuracy.

\begin{algorithm}[H]
\caption{Reptile Meta-Training Loop}
\label{alg:meta_training}
\begin{algorithmic}[1]
\Require Initial meta-parameters $\theta_{0}$, pool of episodic tasks $\mathcal{T}$, validation set $\mathcal{V}$, meta step size $\alpha$, task learning rate $\gamma$, inner epochs $\mathcal{E}$, outer epochs $M$, patience $\mathcal{P}$

\Ensure Best meta-parameters $\theta^{*}$
\State $\theta \gets \theta_{0}$

\For{outer epoch $m=1$ \textbf{to} $\mathcal{M}$}
    \State Sample few-shot task $\mathcal{S}\sim\mathcal{T}$
    \State \textbf{clone} meta-model: $\theta_{0}' \gets \theta$
    \State Initialize task optimizer (learning rate $=\gamma$) and cosine scheduler
    
    \For{inner epoch $\mathcal{e}=1$ \textbf{to} $\mathcal{E}$}
        \State Generate span projections
        \State Compute $\mathcal{L}_{\text{proto}},\;\mathcal{L}_{\text{span}}$
        \State $\mathcal{L} \gets \mathcal{L}_{\text{proto}}+\mathcal{L}_{\text{span}}$
        \State Perform a gradient descent step on $\theta'$
    \EndFor
    
    \State $\theta_{1}' \gets \theta'$  \Comment{fine-tuned parameters}
    
    \State \textbf{Reptile update:}
    \[
        \theta \;\leftarrow\; \theta + \alpha \bigl(\theta_{1}' - \theta_{0}'\bigr)
        \tag*{(Reptile)}
    \]
    \State $\ell_2$-normalize all prototype vectors in $\theta$
    \State Evaluate $F_1$ on $\mathcal{V}$; keep best weights $\theta^{*}$
    
    \If{no $F_1$ improvement for $\mathcal{P}$ outer epochs}
        \State \textbf{break}   \Comment{early stopping}
    \EndIf
    
\EndFor
\State \Return $\theta^{*}$
\end{algorithmic}
\end{algorithm}

\section{Discussion}
\label{sec:discussion}
\subsection{Prototype Representation and Optimization}
\label{ssec:prototype_representation_and_optimization}
We adopt a multi-prototype strategy, allocating $M$ distinct prototypical vectors per category to capture the variability within the same category. We chose cosine similarity over Euclidean distance for two reasons:

\begin{enumerate}[leftmargin=*,itemsep=0.3em]
    \item In high-dimensional spaces, Euclidean distances are less effective metrics~\cite{aggarwal2001surprising}, whereas the cosine distance remains discriminative under $\ell_2$ normalization~\cite{korenius2007principal}.
    
    \item Our projection vectors are already normalized, so angular objectives converge more smoothly and yield more precise decision boundaries.
\end{enumerate}

\begin{figure*}[p]
    \centering
    \includegraphics[height=0.45\textheight]{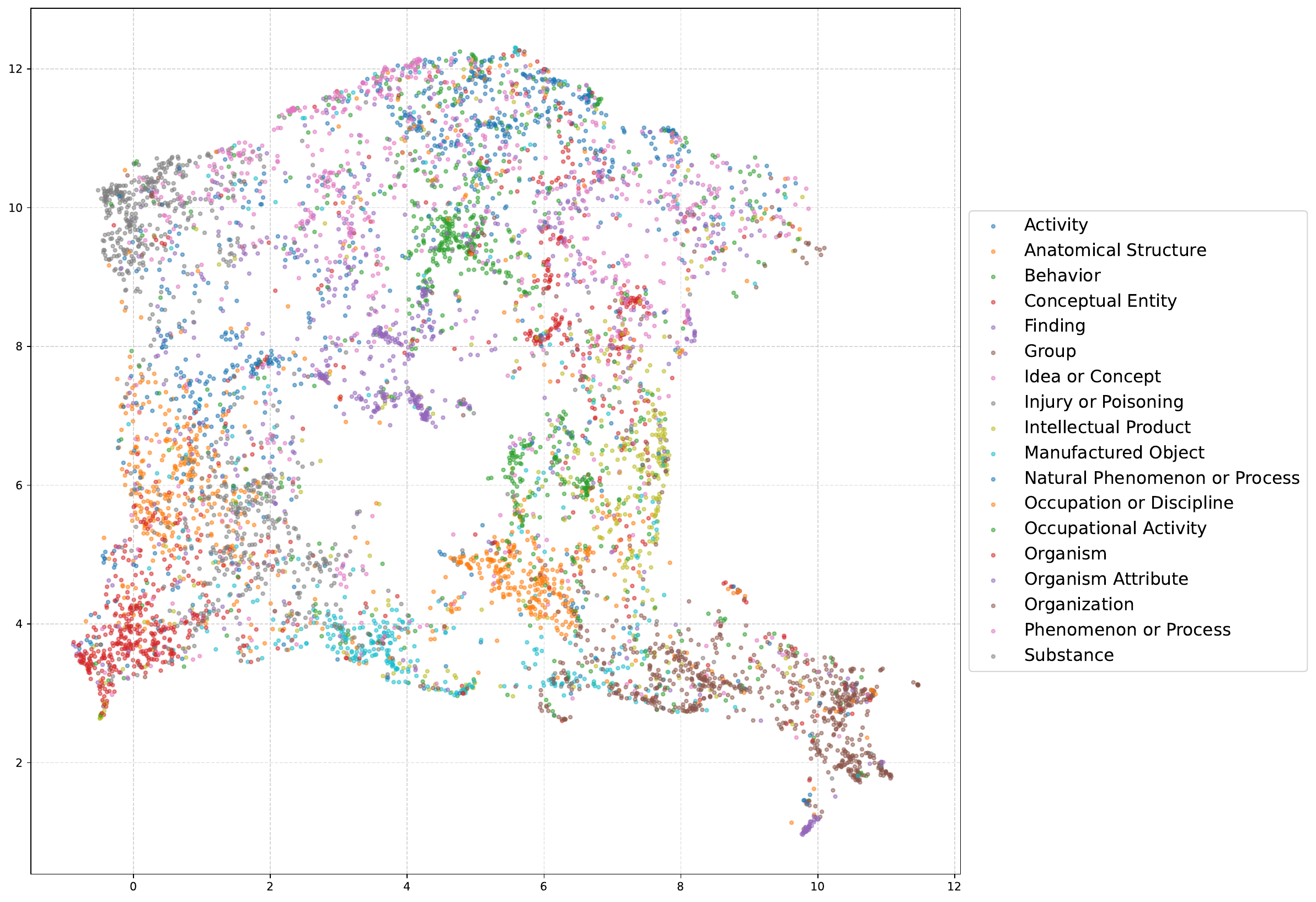}
    \captionsetup{justification=centering}
    \caption{ Two-dimensional UMAP projection of span representations.}
    \Description{A two-dimensional scatter plot showing the UMAP projection of span representations from the 19-way fastText-based NER model, colored by semantic type.}
    \label{fig:umap}
\end{figure*}

Figure~\ref{fig:umap} visualizes the vectors projected from the query set trained with a 0.3 training split ratio of the NER model based on fastText of 19 ways via the UMAP dimensionality reduction~\cite{mcinnes2020umap} condition. The plot indicates that some categories have converged into discrete clusters; however, the number of clusters is less than the number of prototypes assigned per category that we set.

\subsection{Performance Comparison}
\label{ssec:performance_comparison}
We benchmarked two embedding backbones: a lightweight fastText + BiLSTM encoder and a contextualized BERT model. As reported in Table~\ref{tab:perf_summary}, fastText embedding achieved a comparable macro-F\textsubscript{1} performance compared to BERT embedding. Considering its minimal memory footprint, faster inference, and straightforward deployment, the fastText pipeline may be a cost-effective option when working with constrained data and computing resources.

\begin{table}[t]
    \centering
    \small
    \caption{Macro-F\textsubscript{1} (\%) of \textbf{fastText} vs.\ \textbf{BERT} under various training-query split ratios for 19-way and 50-way classification.}
    \label{tab:perf_summary}
    \resizebox{\columnwidth}{!}{%
    \begin{tabular}{@{}lcccccc@{}}
        \toprule
        \textbf{19-way} & 0.8 & 0.7 & 0.6 & 0.5 & 0.4 & 0.3 \\
        \midrule
        fastText & 43.28 & 46.39 & 49.45 & 47.58 & 50.18 & 50.80 \\
        BERT     & 46.37 & 45.97 & 48.26 & 47.68 & 48.09 & 51.03 \\
        \addlinespace
        \textit{Mean} & \multicolumn{6}{c}{fastText = 47.95 \;|\; BERT = 47.90}\\
        \midrule
        \textbf{50-way} & 0.8 & 0.7 & 0.6 & 0.5 & 0.4 & 0.3 \\
        \midrule
        fastText & 46.83 & 50.50 & 49.89 & 51.89 & 50.71 & 46.85 \\
        BERT     & 53.79 & 54.60 & 53.78 & 52.03 & 51.11 & 46.73 \\
        \addlinespace
        \textit{Mean} & \multicolumn{6}{c}{fastText = 49.45 \;|\; BERT = 52.01}\\
        \bottomrule
    \end{tabular}}
\end{table}

\begin{table}[t]
    \centering
    \small
    \caption{Scalability comparison. $\Delta$F$_1$ is the absolute Macro-F\textsubscript{1} drop; “Rel.\ drop” is ${\Delta\text{F}_1}/{\text{F}_1^{\text{small}}}\times100$ (\%). Few-NERD results are from cited papers; MedMentions scores are our 19-way to 50-way experiment.}
    \label{tab:perf_drop_comp}
    \resizebox{\columnwidth}{!}{%
    \begin{tabular}{@{}lcccccc@{}}
        \toprule
        \multirow{2}{*}{\textbf{Method}} &
        \multicolumn{2}{c}{\textbf{Smaller set}} &
        \multicolumn{2}{c}{\textbf{Larger set}} &
        \multirow{2}{*}{$\Delta$F$_1$} &
        \multirow{2}{*}{Rel.\ drop (\%)} \\
        \cmidrule(lr){2-3}\cmidrule(lr){4-5} & Ways & F$_1$ & Ways & F$_1$ & & \\
        \midrule
        NNShot~\cite{yang2020simple} & 5 & 35.74 & 10 & 27.67 & 8.07 & 22.6 \\
        StructShot~\cite{yang2020simple} & 5 & 38.83 & 10 & 26.39 & 12.44 & 32.0 \\
        CONTaiNER & 5  & 53.70 & 10 & 43.87 & 9.83 & 18.3 \\
        Decomposed~\cite{ma2022decomposed} & 5 & 63.23 & 10 & 56.84 & 6.39 & 10.1 \\
        ESD~\cite{wang2022fewshot} & 5 & 52.14 & 10 & 42.15 & 9.99 & 19.2 \\
        SpanProto & 5 & 65.89 & 10 & 59.37 & 6.52 & 9.9 \\
        Three-stage~\cite{ji2024novel} & 5 & 65.22 & 10 & 58.35 & 6.87 & 10.5 \\
        \midrule
        \textbf{Ours} & 19 & 50.80 & 50 & 46.85 & 3.95 & \textbf{7.8} \\
        \bottomrule
    \end{tabular}}
\end{table}

We further assessed robustness under varying data budgets, training in $30\%$ -$80\%$ of available samples. In particular, F\textsubscript{1} remains stable or improves with fewer examples (Table~\ref{tab:perf_summary}), underscoring the data efficiency of our reptile-based few-shot paradigm.

\begin{figure*}[p]
    \centering
    \includegraphics[height=0.45\textheight]{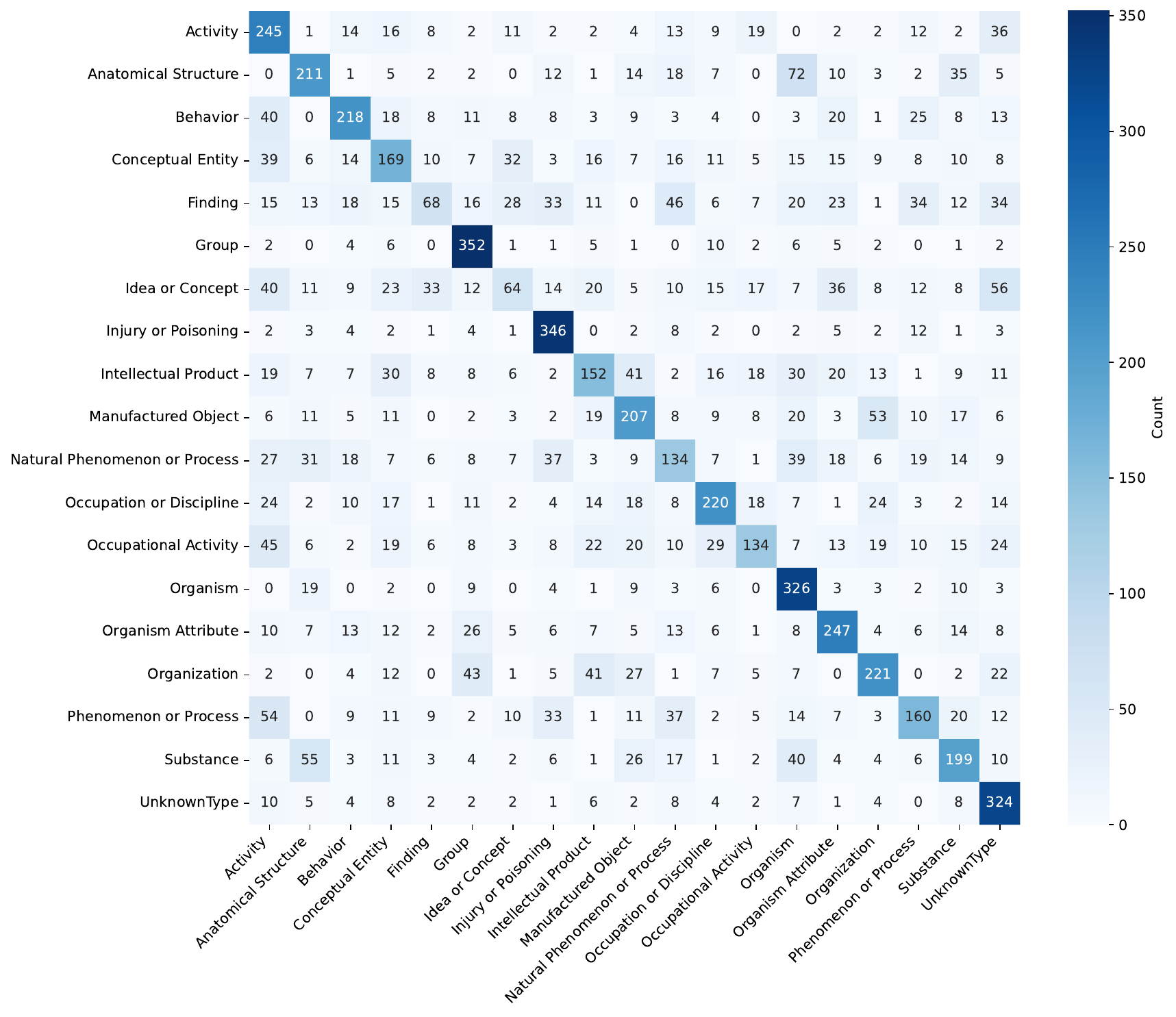}
    \captionsetup{justification=centering}
    \caption{ A confusion matrix of test samples.}
    \Description{A confusion matrix visualizing the classification performance of the 19-way fastText-based NER model on test samples, with rows and columns representing true and predicted semantic types, respectively.}
    \label{fig:confusion_matrix}
\end{figure*}

The confusion matrix in Figure~\ref{fig:confusion_matrix} reveals the performance of one of our models, which uses a token embedding based on fastText + BiLSTM, trained on 30\% of the total samples. This visualization reveals patterns of misclassification that can inform future improvements. The model demonstrates reasonable accuracy, with strong performance in specific categories such as \verb| Group | (352 correct predictions), \verb | Injury or Poisoning | (346), and \verb | Organization | (326), where the recall exceeds 80\%. These categories appear to have more distinct features, possibly due to their semantic uniqueness in the biomedical domain.

Notable confusion patterns include misclassifications between \verb | Anatomical Structure | and \verb | Substance | (35 and 55 instances, respectively), likely due to shared lexical features in biological texts (e.g., tissue or compound). \verb| Phenomenon or Process | and \verb| Natural Phenomenon or Process | show reciprocal errors (37 each), indicating potential label ambiguity in the training data or insufficient examples to learn fine-grained distinctions. The \verb| Activity | is confused with \verb| Occupational Activity | (45) and \verb| Phenomenon or Process | (54), underscoring the model's difficulty in distinguishing between action-oriented and procedural entities, a challenge exacerbated by limited shots in our few-shot setup.

\subsection{Class Scalability Analysis}
\label{ssec:class_scalability_analysis}
\paragraph{Caveat.} Previous work on class scalability has mainly used \emph{Few-NERD}, a general domain benchmark. Therefore, absolute scores are not directly comparable to our biomedical runs. Instead, we focus on the relative performance drop when the number of categories roughly doubles; this ratio is domain independent and serves as a practical proxy for scalability.

Our model loses only \textbf{7.8\%} when the label space expands from 19 to 50 biomedical types, while representative baselines show 10 to 32\% degradation when categories double in Few-NERD (Table~\ref{tab:perf_drop_comp}). We could not train baseline models on MedMentions due to VRAM limitations in Google Colab.

We attribute this robustness to our supervised contrastive objective. By minimizing category-wise minimum distances between spans and multiple prototypes, gradient domination from primary categories is alleviated, yielding balanced decision boundaries even under extreme category imbalance.

\subsection{Ablation Study}
\label{ssec:ablation_study}
To isolate the contribution of each architectural decision, we ablate three key components while keeping all other hyperparameters fixed (fastText + BiLSTM backbone, 0.3 split). Table \ref{tab:ablation_result} investigates the contribution of each architectural or training ingredient to macro-F\textsubscript{1} in general.

\begin{table}[t]
    \centering
    \small
    \caption{Ablation on 19-way task (Macro-F\textsubscript{1}, \%).}
    \label{tab:ablation_result}
    \begin{tabular}{@{}lcc@{}}
        \toprule
        \textbf{Configuration} & \textbf{Change} & \textbf{Macro-F\textsubscript{1}} \\
        \midrule
        Complete model & — & \textbf{50.80} \\
        Single prototype/category & $M=1$ & 49.13 \\
        Hard-neg. Off (random) & disabling hard-neg.\ mining & 56.66 \\
        CE loss & cross-entropy (no contrastive)  & 2.70 \\
        \bottomrule
    \end{tabular}
\end{table}

Replacing the supervised-contrastive objective with vanilla cross entropy causes a collapse to near-random performance (2.7\%). It highlights that the contrastive learning process makes a substantial contribution in a few-shot setting. This property is crucial when there is an extreme imbalance of samples in each category.

Collapsing each category with a single centroid reduces F\textsubscript{1} by approximately 1.7 percentage points. The result confirms our intuition that biomedical entity categories are semantically diverse (e.g., gene symbols vs. full protein names) and therefore benefit from distributed rather than monolithic representations.

Turning off hard-negative sampling and returning to uniformly selected negatives increased performance to 56.66\%. We hypothesize that gradient instability causes this, which is exacerbated by hard-negative mining a handful of high-loss examples, producing steep, noisy gradients that counteract the regularizing effect of contrastive loss. We believe that when the training set is already small, this variance outweighs the potential benefit of focusing on hard-negative samples.

\subsection{Tag Set Extension}
\label{ssec:tag_set_extension}
In addition to validating the model on a fixed set of labels, we experimented with adding previously unseen labels during the training process to assess the model’s ability to retain previously learned information while integrating new information.

From the 19-way label set, we randomly split the 18 labels except for \verb| UnknownType | into six groups of three. For each split, the model was trained for 100 epochs on only two-thirds of the labels not in the split, alongside \verb| UnknownType | in the initial phase (phase 1). We then trained the model for an additional 100 epochs on the complete set of labels (Phase 2). We repeated the process three times with different seeds for each iteration (42, 123, 999).

\begin{itemize}
    \item \textbf{Split A}: Anatomical Structure, Idea or Concept, Injury or Poisoning, Intellectual Product, Occupation or Discipline, Organization
    
    \item \textbf{Split B}: Conceptual Entity, Manufactured Object, Natural Phenomenon or Process, Organism Attribute, Phenomenon or Process, Substance
    
    \item \textbf{Split C}: Activity, Behavior, Finding, Group, Occupational Activity, Organism
\end{itemize}

\begin{table}[t]
    \centering
    \caption{Meta-model F\textsubscript{1} scores across 3 iterations (mean $\pm$ 95\% CI). Full indicates that all categories were used in F\textsubscript{1} calculation, while Base includes only the labels used in the first phase of training.}
    \label{tab:tag_set_extension}
    \begin{tabular}{c c c c}
        \toprule
        \textbf{Split} & \textbf{Phase} & \textbf{Full F\textsubscript{1} (\%)} & \textbf{Base F\textsubscript{1} (\%)} \\
        \midrule
        A & 1 & - & 51.36 [47.74, 54.98] \\
         & 2 & 52.47 [49.25, 55.69] & 56.21 [53.36, 59.06] \\
        B & 1 & - & 55.53 [51.55, 59.51] \\
         & 2 & 52.43 [52.27, 52.60] & 54.30 [53.48, 55.13] \\
        C & 1 & - & 53.59 [50.87, 56.31] \\
         & 2 & 51.06 [48.10, 54.02] & 51.70 [50.55, 52.86] \\
        \bottomrule
    \end{tabular}
\end{table}

Table~\ref{tab:tag_set_extension} describes the performance on the Base subset of labels, which remains similar or improved despite the addition of previously unseen labels. F\textsubscript{1} in Split A showed a statistically significant performance improvement ($p=0.039$) in the categories of the Base subset. Split B and C showed a performance decrease, but the effects were minimal ($p=0.554$ for split B, $p=0.211$ for split C). These results demonstrate the robustness of our model to changes in the distribution of labels.

Comparison of the full F\textsubscript{1} and the base F\textsubscript{1} score for each split showed that new labels can be added late in the training process without a decrease in performance ($p=0.653$ for split A, $p=0.128$ for split B, $p=0.218$ for split C).

\section{Conclusion}
\label{sec:conclusion}
We proposed ReProCon, a cosine-contrastive few-shot NER framework for biomedical text that integrates supervised contrastive learning with Reptile meta-updates. Using a lightweight \textbf{fastText + BiLSTM} encoder, ReProCon achieves $\sim$99\% of a baseline Macro-F\textsubscript{1} based on BERT while requiring significantly less memory. Performance remains stable with a label budget of 30\% and decreases by only 3.95 percentage points (7.8\%) when expanding from 19 to 50 categories. These results demonstrate that state-of-the-art biomedical few-shot NER is achievable without large language models, making ReProCon well-suited for resource-constrained environments.

In future work, we plan to:
\begin{enumerate}[leftmargin=*,itemsep=0.3em]
    \item Conduct cross-domain meta-training from PubMed abstracts to clinical notes and patent texts to evaluate domain-shift robustness.
    
    \item Comparison of performances between models from other studies, by training on biomedical datasets.
\end{enumerate}

%% The acknowledgments section is defined using the "acks" environment (and NOT an unnumbered section). It ensures the proper identification of the section in the article metadata and the consistent spelling of the heading.
\begin{acks}
We acknowledge the Discord user \verb| aneesh_qai |, who suggested valuable ideas, including Xavier uniform initialization for the prototype tensor, cosine decay scheduling, and mean-max pooling for span representation using BERT.
\end{acks}

%% The following two lines define the bibliography style and the bibliography file.
% \bibliographystyle{ACM-Reference-Format}
\bibliographystyle{unsrtnat}
\bibliography{myrefs}

\clearpage
\onecolumn
% \appendix{Appendix}
\section*{Appendix}

\begin{table}[ht]
\centering
\caption{50 Categories and Frequencies of Merged Categories.}
\label{tab:50categories}
\begin{tabular}{|l r|l r|}
\hline
Type & Frequency & Type & Frequency \\
\hline
UnknownType & 6,947,708 & Phenomenon or Process & 2,239 \\
Biologic Function & 47,840 & Injury or Poisoning & 2,215 \\
Chemical & 39,839 & Sign or Symptom & 2,104 \\
Qualitative Concept & 33,479 & Clinical Attribute & 2,001 \\
Health Care Activity & 28,221 & Professional or Occupational Group & 1,984 \\
Functional Concept & 26,191 & Substance & 1,910 \\
Fully Formed Anatomical Structure & 23,592 & Food & 1,538 \\
Quantitative Concept & 18,252 & Classification & 1,449 \\
Finding & 16,521 & Body Substance & 1,403 \\
Spatial Concept & 14,354 & Organism & 1,234 \\
Research Activity & 10,644 & Individual Behavior & 1,127 \\
Idea or Concept & 10,296 & Biomedical Occupation or Discipline & 1,019 \\
Intellectual Product & 10,167 & Social Behavior & 1,015 \\
Temporal Concept & 10,159 & Laboratory or Test Result & 987 \\
Group & 10,066 & Occupational Activity & 969 \\
Eukaryote & 9,288 & Daily or Recreational Activity & 944 \\
Activity & 8,643 & Anatomical Abnormality & 868 \\
Population Group & 6,738 & Family Group & 863 \\
Manufactured Object & 4,082 & Educational Activity & 643 \\
Organism Attribute & 3,864 & Occupation or Discipline & 614 \\
Conceptual Entity & 3,413 & Health Care Related Organization & 554 \\
Natural Phenomenon or Process & 3,162 & Human-caused Phenomenon or Process & 509 \\
Medical Device & 2,638 & Anatomical Structure & 445 \\
Bacterium & 2,243 & Organization & 369 \\
Virus & 1,412 & Governmental or Regulatory Activity & 193 \\
\hline
\end{tabular}
\end{table}

\begin{table}[ht]
\centering
\caption{19 Categories and Frequencies of Merged Categories.}
\label{tab:19categories}
\begin{tabular}{|l r|l r|}
\hline
Type & Frequency & Type & Frequency \\
\hline
UnknownType & 6,947,708 & Organism & 14,160 \\
Idea or Concept & 112,603 & Intellectual Product & 11,092 \\
Natural Phenomenon or Process & 51,014 & Activity & 9,044 \\
Substance & 44,679 & Manufactured Object & 6,955 \\
Occupational Activity & 40,945 & Organism Attribute & 5,865 \\
Anatomical Structure & 24,858 & Conceptual Entity & 3,413 \\
Group & 19,616 & Phenomenon or Process & 2,744 \\
Finding & 19,594 & Behavior & 2,670 \\
Injury or Poisoning & 2,215 & Occupation or Discipline & 1,633 \\
Organization & 923 &  &  \\
\hline
\end{tabular}
\end{table}

\clearpage

\end{document}